\documentclass[11pt]{article-mod}

\usepackage[american,english]{babel}

\usepackage{epsfig}
\usepackage{theorem}
\usepackage{enumerate}
\usepackage{apa-uiuc}
\usepackage{amsmath}
\usepackage{amssymb}
\usepackage{boxedminipage}
\usepackage{subfigure}
\usepackage{url}

\begin{document}
\begin{titlepage}
\oddsidemargin 6mm
\vspace*{2.0in}
\vspace*{3ex}
\begin{center}
{\bf Multiobjective hBOA, Clustering, \\and Scalability}
 
\addvspace{0.3in}
{\bf Martin Pelikan\\Kumara Sastry\\David E. Goldberg}\\
\addvspace{0.45in}
IlliGAL Report No. 2005005 \\
February 2005 \\
 
\vspace*{2.8in}
Illinois Genetic Algorithms Laboratory \\
University of Illinois at Urbana-Champaign \\
117 Transportation Building \\
104 S. Mathews Avenue
Urbana, IL 61801 \\
Office:  (217) 333-2346\\
Fax: (217) 244-5705 \\
\end{center}
\end{titlepage}
 
\title{\bf Multiobjective hBOA, Clustering, and Scalability}

\author{
{\bf Martin Pelikan}\\
Dept. of Mathematics and Computer Science, 320 CCB\\
University of Missouri at St. Louis\\
8001 Natural Bridge Rd.\\
St. Louis, MO 63121\\
\url{pelikan@cs.umsl.edu}
\and
{\bf Kumara Sastry}\\
Illinois Genetic Algorithms Laboratory, 107 TB\\
University of Illinois at Urbana-Champaign\\
104 S. Mathews Ave.\\
Urbana, IL 61801\\
\url{kumara@illigal.ge.uiuc.edu}
\and 
{\bf David E. Goldberg}\\
Illinois Genetic Algorithms Laboratory, 107 TB\\
University of Illinois at Urbana-Champaign\\
104 S. Mathews Ave.\\
Urbana, IL 61801\\
\url{deg@illigal.ge.uiuc.edu}}

\maketitle 


\begin{abstract}
\begin{sloppy}
This paper describes a scalable algorithm for solving multiobjective decomposable problems by combining the hierarchical Bayesian optimization algorithm (hBOA) with the nondominated sorting genetic algorithm (NSGA-II) and clustering in the objective space. It is first argued that for good scalability, clustering or some other form of niching in the objective space is necessary and the size of each niche should be approximately equal. Multiobjective hBOA (mohBOA) is then described that combines hBOA, NSGA-II and clustering in the objective space. The algorithm mohBOA differs from the multiobjective variants of BOA and hBOA proposed in the past by including clustering in the objective space and allocating an approximately equally sized portion of the population to each cluster. The algorithm mohBOA is shown to scale up well on a number of problems on which standard multiobjective evolutionary algorithms perform poorly. 
\end{sloppy}
\end{abstract}

\noindent
{\bf Keywords}\\
Genetic algorithms, estimation of distribution algorithms, multiobjective optimization, BOA, nondominated sorting, NSGA-II, clustering.


\section{Introduction}
One of the important strengths of evolutionary algorithms is that they can deal with multiple objectives and find Pareto-optimal solutions, which define a tradeoff between these objectives. The Pareto-optimal front can be exploited to select solutions appropriate for each particular application without having to weigh the objectives in advance or reduce the multiple objectives in some other way. A number of multiobjective evolutionary algorithms were proposed in the past, including the nondominated sorting genetic algorithm II (NSGA-II)~\cite{Deb:02}, the improved strength Pareto evolutionary algorithm (SPEA2)~\cite{Zitzler:01}, the Pareto-archived evolution strategy (PAES)~\cite{Knowles:99}, the multiobjective genetic algorithm (MOGA)~\cite{Fonseca:93a}, the niched-Pareto genetic algorithm (NPGA)~\cite{Horn:94}, and the multiobjective Bayesian optimization algorithm (mBOA)~\cite{Khan:03,Khan:02,Laumanns:02}. 

\begin{sloppy}
However, all studies of multiobjective evolutionary algorithms focused either on whether or not an algorithm could discover a wide and dense Pareto-optimal front or on practical applications, but they overlook {\em algorithm scalability}, that means, how the time complexity of a multiobjective evolutionary algorithm grows with problem size. Recent work~\cite{Sastry:05*} has shown that although the proposed multiobjective evolutionary algorithms can provide a high-quality Pareto-optimal front on isolated problem instances of relatively small size, the time complexity of these algorithms can often grow prohibitively fast and the algorithms thus do not scale up well.
\end{sloppy}

The purpose of this paper is to present a scalable multiobjective evolutionary algorithm that can solve decomposable multiobjective problems in low-order polynomial time. The algorithm consists of three main ingredients: (1) Model-building, model-sampling, and replacement procedures of hBOA \cite{Pelikan:01*,Pelikan:03b,Pelikan:book}, (2) nondominated sorting and crowding-distance assignment of NSGA-II~\cite{Deb:02}, and (3)~clustering in the objective space~\cite{Thierens:01}.

The paper starts with an overview of prior work on multiobjective estimation of distribution algorithms (EDAs) and an introduction to basic concepts used in mohBOA. Section~\ref{section-mohBOA} describes mohBOA. Section~\ref{section-experiments} describes experiments and presents experimental results. 
Finally, Section~\ref{section-conclusions} summarizes and concludes the paper.  


\section{Background}
\label{section-background}
This section starts by summarizing prior work on multiobjective estimation of distribution algorithms (EDAs). The section then discusses basic components of the multiobjective hierarchical Bayesian optimization algorithm (mohBOA): (1) The hierarchical Bayesian optimization algorithm (hBOA)~\cite{Pelikan:99a,Pelikan:book}, (2) the nondominated sorting genetic algorithm (NSGA-II)~\cite{Deb:02}, and (3) the k-means algorithm~\cite{MacQueen:67} for clustering in the objective space~\cite{Thierens:01}. 

\subsection{Prior work on multiobjective EDAs}
Estimation of distribution algorithms (EDAs)~\cite{Muhlenbein:96**,Pelikan:02,Larranaga:02}, also called probabilistic model-building genetic algorithms (PMBGAs)~\cite{Pelikan:02} and iterated density estimation algorithms (IDEAs)~\cite{Bosman:00*}, replace standard variation operators such as crossover and mutation by building a probabilistic model of selected candidate solutions and sampling the built model to generate new solutions. Several multiobjective EDAs were proposed in the past for different variants of BOA~\cite{Khan:02,Khan:03,Laumanns:02} and for EDAs with mixtures of univariate and tree models~\cite{Thierens:01}.

Khan~\cite{Khan:02,Khan:03} proposed multiobjective BOA (mBOA) and multiobjective hBOA (mhBOA) by combining BOA and hBOA with the selection and replacement mechanisms of NSGA-II~\cite{Deb:02}. Tests on challenging decomposable multiobjective problems indicated that without identifying and exploiting interactions between different string positions, some decomposable problems become intractable using standard variation operators (crossover and mutation). On the other hand, mBOA and mhBOA could solve decomposable and hierarchical problems relatively efficiently. 

\citeN{Laumanns:02} combined mixed BOA~\cite{Ocenasek:02} with the selection and replacement mechanisms of SPEA2~\cite{Zitzler:01}. Since mixed BOA can be applied to problems with both discrete and continuous variables, the resulting algorithm can also be applied to vectors over both types of variables. The algorithm was tested on knapsack where it was shown to dominate NSGA-II, SPEA, and SPEA2 in most instances. 

\citeN{Ahn:thesis} combined real-coded BOA~\cite{Ahn:04} with the 
selection procedure of NSGA-II with a sharing intensity measure and 
modified NSGA-II crowding mechanism. 

Incorporating learning and sampling of multivariate probabilistic models was an important step toward competent multiobjective solvers, because it allowed standard multiobjective genetic algorithms, such as NSGA-II and SPEA2, to solve problems that necessitate some form of linkage learning. However, the primary purpose of all presented tests was to examine good coverage of the Pareto-optimal front but they overlooked scalability. It was later indicated~\cite{Sastry:05*} that the proposed multiobjective EDAs do not scale up well on some decomposable problems without some form of clustering as is discussed in this paper, and neither do standard multiobjective GAs, such as NSGA-II.

\citeN{Thierens:01} combined simple EDAs with univariate and tree models with nondominated tournament selection and clustering. Clustering was used to split the population into subpopulations where a separate model is built for each subpopulation. The use of clustering was yet another important step toward scalable EDAs and other multiobjective evolutionary algorithms. The number of samples generated from the model of each cluster was set to encourage the sampling of extreme solutions and that is why the methods described in~\citeN{Thierens:01} use clustering in a similar way as we use it in this paper. However, here we use clustering to explicitly ensure that each part of the Pareto-optimal front will have sufficiently many candidates in the population.

The following section describes the hierarchical BOA. Next, multiobjective optimization is discussed. Finally, NSGA-II and the k-means clustering algorithm are described.

\subsection{Hierarchical BOA (hBOA)}
The hierarchical Bayesian optimization algorithm (hBOA) \cite{Pelikan:book,Pelikan:99a} evolves a population of candidate solutions to the given problem. The first population is usually generated at random. The population is updated for a number of iterations using two basic operators: (1) selection and (2) variation. The selection operator selects better solutions at the expense of the worse ones from the current population, yielding a population of promising candidates. The variation operator starts by learning a probabilistic model of the selected solutions that encodes features of these promising solutions and the inherent regularities. hBOA uses Bayesian networks with local structures~\cite{Chickering:97,Friedman:99} to model promising solutions. The variation operator then proceeds by sampling the probabilistic model to generate new candidate solutions. The new solutions are incorporated into the original population using the restricted tournament replacement (RTR)~\cite{Harik:95a}, which ensures that useful diversity in the population is maintained for long periods of time. A more detailed description of hBOA can be found in~\citeN{Pelikan:book} and~\shortciteN{Pelikan:99a}. 


\subsection{Multiobjective optimization}
\begin{figure}[t]
\centering
\epsfig{file=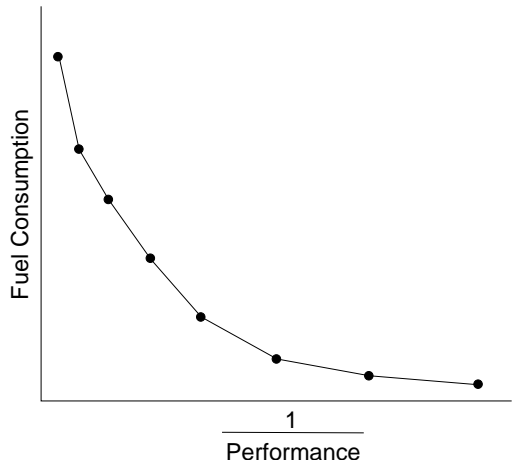,width=0.4 \textwidth}
\caption{Pareto-optimal front in two-objective engine design.}
\label{fig-engine}
\end{figure}

In multiobjective optimization, the task is to find a solution or solutions that are optimal with respect to multiple objectives. For example, one may want optimize the design of an engine to both maximize its performance as well as minimize its fuel consumption. There are two basic approaches to solving multiobjective optimization problems: (1) weigh the objectives in some way, yielding a single-objective problem where the objective consists of a weighted sum of all objectives, and (2) find the Pareto-optimal front, which is defined as the set of solutions that can be improved with respect to any objective only at the expense of their quality with respect to at least one other objective; for example, the performance of an engine design on the Pareto-optimal front could be improved only at the expense of its fuel consumption and the other way around (see Figure~\ref{fig-engine}).

Pareto optimality can be easily explained using the concept of {\em dominance}. We say that a candidate solution $A$ dominates a candidate solution $B$ if $A$ is better than $B$ with respect to at least one objective but A is not worse than $B$ with respect to all other objectives. For example, engine $A$ dominates engine $B$ if $A$ is better than $B$ with respect to both performance as well as fuel consumption. The Pareto-optimal front is then a subset of all candidate solutions that are not dominated by any other candidate solution.

The primary advantage of finding the Pareto-optimal front as opposed to finding the optimum to a single-objective problem created by weighing the objectives is that sometimes it is difficult or impossible to weigh the objectives appropriately to find satisfactory solutions. Furthermore, finding the Pareto-optimal front reveals the relationship among the objectives, which can be used to decide which of the solutions on this front is best for each particular problem instance.  In this paper we focus on the discovery of the Pareto-optimal front because the application of hBOA or other advanced evolutionary algorithms to any single-objective problem is straightforward. The task is to find solutions across the entire Pareto-optimal front and cover the front as well as possible.

A great overview of multiobjective evolutionary algorithms and their comparison can be found in~\citeN{Deb:01} and~\shortciteN{Coello:01}.


\subsection{Nondominated sorting GA (NSGA-II)}
The nondominated sorting genetic algorithm (NSGA-II) modifies selection and replacement of standard genetic algorithms to enable the discovery of a wide-spread, dense Pareto-optimal front. Here we briefly describe the basic principle of the selection and replacement operators of NSGA-II.

The selection operator in NSGA-II starts by partially sorting the population using dominance. First, rank $1$ is assigned to the subset of the current population that consists of solutions that are not dominated by any solution in the population. Next, solutions that are not dominated by any of the remaining, unranked solutions are selected and given rank $2$. The process of ranking solutions continues by always considering solutions that are not dominated by the remaining solutions and assigning increasing ranks to these solutions. In this manner, the solutions that are dominated by least solutions are given lower ranks than the solutions that are dominated by most solutions. With respect to Pareto optimality, solutions with lower ranks should be given priority. 

In addition to the ranking, each candidate solution is assigned a crowding distance, which estimates how dense the current Pareto-optimal front is in the vicinity of this solution. The higher the crowding distance of a solution, the more isolated the solution. The crowding distance is computed for each rank separately. The candidate solutions are first sorted according to each objective and the crowding distance of each solution is computed by considering the distance of its nearest neighbors in this ordering~\cite{Deb:02}. See Figure~\ref{fig-crowding-distance-assignment} for the pseudocode of the crowding distance assignment algorithm.

\begin{figure}[t]
\begin{verbatim}
crowding-distance-assignment(P)
   for each rank r (nondominated sorting)
      P' = select solutions with rank r from P;
      N = size(P');
      for each solution X in P'
          d(X)=0;
      for each objective m
          Q = sort P' using m-th objective;
          d(Q(1))=infinity;
          d(Q(N))=infinity;
          for i=2 to N-1
             d(Q(i))=d(Q(i))+Q(i+1).m-Q(i-1).m;
   return d;
\end{verbatim}
\caption{Crowding distance assignment in NSGA-II. For a solution {\tt X}, {\tt X.m} denotes the value of {\tt m}th objective for {\tt X}. {\tt Q(i)} denotes {\tt i}th candidate solution in population {\tt Q}.}
\label{fig-crowding-distance-assignment}
\end{figure}

To compare quality of two solutions, their ranks are compared first. If the ranks of the solutions differ, the solution with the lower rank is better. If the ranks of both the solutions are equal, the solution with a greater crowding distance wins. If both the ranks as well as the crowding distances are equal, the winner is determined randomly. A pseudocode for the comparison of two solutions is shown in Figure~\ref{fig-nondominated-selection}. This comparison procedure can be used in any standard selection operator, such as tournament or truncation selection. In both the original NSGA-II as well as mohBOA, tournament selection is used. 

\begin{figure}[t]
\begin{verbatim}
compare(A,B)
   if (rank(A)<rank(B)) then better(A,B)=A;
   if (rank(A)>rank(B)) then better(A,B)=B;
   if (rank(A)=rank(B))
      then if (crowding(A)>crowding(B)) 
              then better(A,B)=A;
           if (crowding(A)<crowding(B)) 
              then better(A,B)=B;
           if (crowding(A)=crowding(B)) 
              then better(A,B)=random(A,B);
\end{verbatim}
\caption{Nondominated crowding selection in NSGA-II.}
\label{fig-nondominated-selection}
\end{figure}

The selected population of solutions undergoes mutation and crossover and is combined with the original population to form the new population of candidate solutions. NSGA-II uses an elitist replacement mechanism to combine the parent population $P(t)$ and the offspring population $O(t)$ to form the new population $P(t+1)$. The replacement operator in NSGA-II starts by merging the two populations $P(t)$ and $O(t)$ into one population. Ranks and crowding distances are then computed for all solutions in the merged population, and the nondominated crowding comparison operator is used to select best solutions from the merged population. The best solutions are then transferred to the new population $P(t+1)$. For more details about NSGA-II, see~\shortciteN{Deb:02}.

\subsection{K-means clustering in the objective space}
Given a set $X$ of $N$ points, k-means clustering~\cite{MacQueen:67} splits $X$ into $k$ clusters or subsets with approximately same variance. The algorithm proceeds by updating a set of $k$ cluster centers where each center defines one cluster. The cluster centers can be initialized randomly but more advanced algorithms can also be used to initialize the centers. 

Each iteration consists of two steps. In the first step, each point in $X$ is attached to the closest center (ties can be resolved arbitrarily). In the second step, cluster centers are recomputed so that each center is the center of mass of the points attached to it. The algorithm terminates when all points in $X$ remain in the same cluster after recomputing cluster centers and reassigning the points to the newly computed centers. Points attached to each cluster center define one cluster. The numbers of points in different clusters can differ significantly if points in $X$ are not distributed uniformly and some clusters may even become empty. Sometimes it is necessary to rerun k-means several times and use the result of the best run. 

\begin{sloppy}
In decomposable multiobjective problems where the objectives compete in a number of problem partitions, using traditional selection and replacement mechanisms necessitates exponentially scaled populations to discover the entire Pareto-optimal front~\cite{Sastry:05*}. The reason for this behavior is that the niches on the extremes of the Pareto-optimal front (maximizing most partitions with respect to one particular objective) can be expected to be exponentially smaller than the niches in the middle~\cite{Sastry:05*}. To alleviate this problem, it is necessary to process different parts of the Pareto-optimal front separately and allocate a sufficiently large portion of the population to each part of the Pareto-optimal front. 
\end{sloppy}

It is important to note that other algorithms, such as NSGA-II and SPEA2, also include mechanisms that attempt to deal with a good coverage of a wide Pareto-optimal front. However, these mechanisms are insufficient for some decomposable multiobjective problems because they result in creating exponentially large niches in the middle of the Pareto-optimal front while eliminating extremes. This leads to poor scalability, which is supported with experimental results shown in Section~\ref{section-experiments}.

Allocating comparable space to each part of the Pareto-optimal front can be ensured by using clustering in the objective space (the space of fitness values) as suggested in~\citeN{Thierens:01}. $X$ thus consists of $m$-dimensional vectors of real numbers where $m$ is the number of objectives. To reduce the number of iterations until the creation of reasonable clusters, the cluster centers can be initialized by ordering points according to one objective and assigning the $i$th center to $(N/(2k) + i [N/k])$th point in this ordering. By forcing each cluster to produce an equal number of new candidate solutions (using an appropriate variation operator), regular coverage of the Pareto-optimal front can be ensured even for difficult decomposable multiobjective problems. 


\section{Multiobjective hBOA (mohBOA)}
\label{section-mohBOA}
This section describes the multiobjective hBOA (mohBOA), which combines hBOA, NSGA-II, and clustering in the objective space. 

The pseudocode of mohBOA is shown in Figure~\ref{figure-mohBOA}. Like hBOA, mohBOA generates the initial population of candidate solutions at random. The population is first evaluated. Similarly as in other evolutionary algorithms, each iteration starts with selection. However, instead of using standard selection methods, mohBOA first uses the nondominated crowding of NSGA-II to rank candidate solutions and assign their crowding distances. The ranks and crowding distances then serve as the basis for applying standard selection operators. For example, binary tournament selection can then be used where the winner of each tournament is determined by the ranks and crowding distances obtained from the nondominated crowding. 

\begin{figure}[t]
\begin{verbatim}
multiobjective-hBOA(N, k, objectives)
  t := 0;
  generate initial population P(0) of size N;
  evaluate P(0);
  while (not done) {
    rank members of P(t) using nondom. crowding;
    select S(t) from P(t) based on the ranking;
    cluster S(t) into k clusters;
    build Bayesian network with local structures 
       for each cluster;
    create O(t) of size N by sampling the model 
       for each cluster to generate N/k solutions;
    evaluate O(t);
    combine O(t) and P(t) to create P(t+1);
    t:=t+1;
  }
\end{verbatim}
\caption{Pseudocode of the multiobjective hBOA.}
\label{figure-mohBOA}
\end{figure}

After selecting the population of promising solutions, k-means clustering is applied to this population to obtain a specified number of clusters. Usually, some clusters remain empty and are thus not considered in the recombination phase. A separate probabilistic model is built for each cluster and used to generate a part of the offspring population. To encourage an equal coverage of the entire Pareto-optimal front, the model for each cluster is used to generate the same number of new candidate solutions. 

The population consisting of all newly generated solutions is then combined with the original population to create the new population of candidate solutions. We use two methods to combine the two populations: (1) the elitist replacement based on the nondominated crowding of NSGA-II, and (2) the restricted tournament replacement (RTR)~\cite{Harik:95a} based on the nondominated crowding.

RTR incorporates the offspring population into the original population one solution at a time. For each offspring solution $X$, a random subset of candidate solutions in the original population is first selected. The solution that is closest to $X$ is then selected from this subset. Here we measure the distance of two solutions by counting the number of bits where solutions differ but it might be advantageous to consider a distance metric defined in the objective space similarly as in k-means clustering; however, using the objective function values to compute distance does not seem to improve results significantly as is shortly verified with experiments. $X$ replaces the selected solution if it is better according to the NSGA-II comparison procedure (Figure~\ref{fig-nondominated-selection}).


\section{Experiments}
\label{section-experiments}
This section describes test problems used in the experiments and presents experimental results. 


\subsection{Test problems}
There were five primary objectives in the design of test problems used in this paper: 
\begin{enumerate}[(i)]
\item  {\bf Scalability.} Test problems should be scalable, that is, it should be possible to increase problem size without affecting the inherent problem difficulty severely.
\item  {\bf Decomposability.} Objective functions should be decomposable.
\item  {\bf Known solution.} Test problems should have a known Pareto-optimal front in order to be able to verify the results.
\item  {\bf Much competition.} The objectives should compete in all or most partitions of an appropriate problem decomposition.
\item  {\bf Linkage learning.} Some test problems should require the optimizer to be capable of linkage learning, that is, of identifying and exploiting interactions between decision variables. 
\end{enumerate}

Two two-objective test problems were used: onemax vs. zeromax and trap-5 vs. inverse trap-5 problem. Both functions assume binary-string representation. This section describes these problems and discusses their difficulty.

\subsubsection{Problem 1: Onemax-zeromax}

\begin{figure}[t]
\centering
\epsfig{file=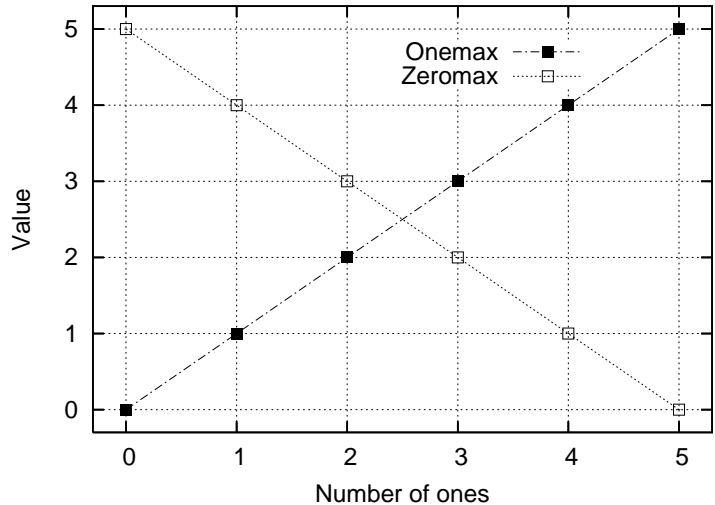,width=0.6\textwidth}
\caption{Onemax and zeromax for a 5-bit string.}
\label{figure-onemax}
\end{figure}

The first test problem consists of two objectives: (1) onemax and (2) zeromax. Onemax is defined as the sum of bits in the input binary string $X=(X_1, X_2, \ldots, X_n)$:
\begin{equation}
onemax(X) = \sum_{i=1}^n X_i
\end{equation}
The task is to maximize the function and thus the optimum of onemax is in the string of all ones. See Figure~\ref{figure-onemax} to visualize onemax for 5-bit strings. Zeromax is defined as the number of positions containing a $0$:
\begin{equation}
zeromax(X) = n-onemax(X)
\end{equation}
The task is to maximize the function and thus the optimum of zeromax is in the string of all zeros. See Figure~\ref{figure-onemax} to visualize zeromax for 5-bit strings. 

Onemax and zeromax are conflicting objectives; in fact, {\em any} modification that increases one objective decreases the other objective. In onemax-zeromax, any binary string is located on the Pareto-optimal front.

\subsubsection{Problem 2: Trap5-invtrap5}
\begin{figure}[t]
\centering
\epsfig{file=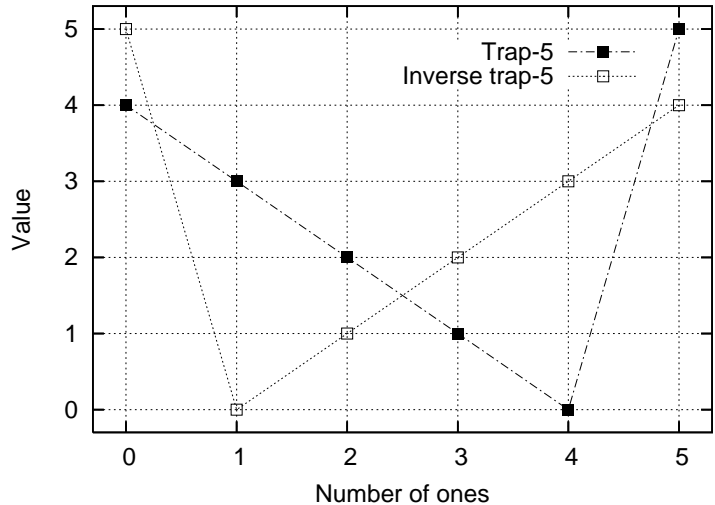,width=0.6\textwidth}
\caption{Trap-5 and inverse trap-5 for one 5-bit block.}
\label{figure-trap5}
\end{figure}

The second test problem consists of two objectives: (1) trap-5 and (2) inverse trap-5. String positions are first (before running the optimizer) divided into disjoint subsets or partitions of 5 bits each. The partitioning is fixed during the entire optimization run, but the algorithm is not given information about the partitioning in advance. Bits in each partition contribute to trap-5 using a fully deceptive 5-bit trap function~\cite{Ackley:87b,Deb:91c} defined as
\begin{equation}
trap_5(u) = 
\left\{
\begin{array}{ll}
5 & \mbox{~~~~~if $u=5$}\\
4-u & \mbox{~~~~~if $u<5$}
\end{array}
\right.
\end{equation}
where $u$ is the number of ones in the input string of 5 bits. The task is to maximize the function and thus the optimum of trap-5 is in the string of all ones. See Figure~\ref{figure-trap5} to visualize trap-5 for one block of 5 bits.

Trap-5 deceives the algorithm away from the optimum if interactions between the bits in each partition are not considered~\cite{Thierens:95,Bosman:99,Pelikan:99a}. That is why standard crossover operators---such as uniform, one-point, and two-point crossover---fail to solve trap-5 unless the bits in each partition are located close to each other in the chosen representation; in fact, standard crossover operators require exponentially scaled population sizes to solve trap-5~\cite{Thierens:95}. Mutation operators require $O(n^5 \log n)$ evaluations to solve trap-5 and, therefore, are also highly inefficient in solving trap-5. 

Inverse trap-5 is defined using the same partitions as trap-5, but the basis function, which is applied to each partition, is modified as follows:
\begin{equation}
invtrap_5(u) = 
\left\{
\begin{array}{ll}
5 & \mbox{~~~~~if $u=0$}\\
u-1 & \mbox{~~~~~if $u>0$}
\end{array}
\right.
\end{equation}

\begin{figure}[t]
\centering
\subfigure[UMDA on onemax-zeromax (log. scale)]{\epsfig{file=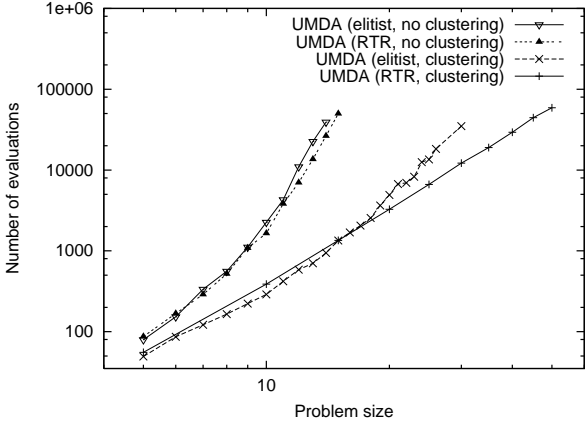,width=0.49 \textwidth}}
\subfigure[GA on onemax-zeromax (log. scale)]{\epsfig{file=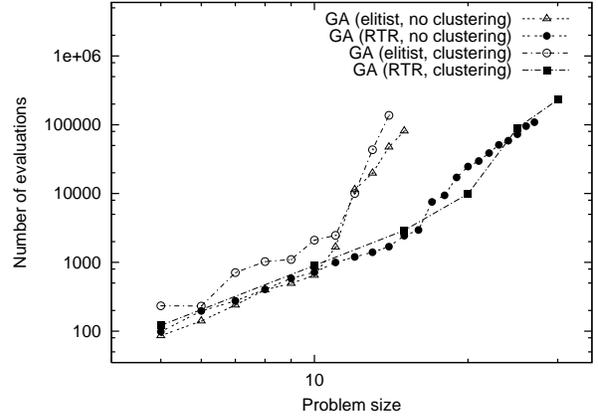, width=0.49 \textwidth}}
\caption{Results on onemax-zeromax indicate that k-means clustering in the objective space leads to a dramatic improvement in performance for both UMDA and GA. Furthermore, they indicate that RTR performs better than the elitist replacement of NSGA-II and that multiobjective UMDA with RTR is capable of solving onemax-zeromax in low-order polynomial time.}
\label{fig-results-onemax}
\end{figure}

The task is to maximize the function and thus the optimum of inverse trap-5 is located in the string of all zeros. See Figure~\ref{figure-trap5} to visualize trap-5 for one block of 5 bits. Inverse trap-5 also deceives the algorithm away from the optimum if the interactions between the bits in each partition are not considered. 

Trap-5 and inverse trap-5 are conflicting objectives. Any solution that sets the bits in each partition either to 0s or to 1s is Pareto-optimal and thus there are $2^{n/5}$ Pareto-optimal solutions. 
%

\subsection{Experimental methodology}
Three recombination operators were tested: 
\begin{enumerate}[(i)]
\item  UMDA recombination~\cite{Muhlenbein:96**} where a probabilistic model with no interactions is used to model and sample solutions, 
\item  two-point crossover and bit-flip mutation, and 
\item  mohBOA recombination based on Bayesian networks with local structures. 
\end{enumerate}
 
For each recombination operator, both aforementioned replacement mechanisms were used (elitist replacement and RTR). 

For all test problems and all algorithms, different problem sizes were examined to study {\em scalability}. For each problem type, problem size and algorithm, bisection was used to determine a minimum population size to find one representative solution for each point on the Pareto-optimal front (solutions with the same values of both objectives are considered equivalent) in 10 out of 10 independent runs. The Pareto-optimal front for an $n$-bit onemax-zeromax consists of $(n+1)$ solutions with unique values of the two objectives whereas the Pareto-optimal front for an $n$-bit trap5-invtrap5 consists of $(n/5+1)$ solutions with unique objective function values. To reduce noise, the bisection method was ran 10 times. Thus, the results for each problem type, problem size, and algorithm correspond to 100 successful runs. Algorithm performance was measured by the number of evaluations until the Pareto-optimal front was completely covered.

The number of generations for UMDA and mohBOA recombination was upper-bounded by $5n$ where $n$ is the problem size (the number of bits), whereas the runs with standard crossover and mutation were given at most $10n$ or $20n$ generations (depending on the test) because of their slower convergence. For GAs, the probability of crossover was $p_c=0.6$, whereas the probability of flipping each bit by mutation was $p_m=1/n$.

To focus only on the effects of different recombination and replacement strategies, the number of clusters in k-means clustering was set to the number of unique solutions on the final Pareto-optimal front (again, solutions with equal objective values are considered equivalent). If the number of clusters cannot be approximated in advance, it can be obtained automatically using for example the Bayesian information criterion (BIC)~\cite{Schwarz78a}.

\subsection{Results}
Figure~\ref{fig-results-onemax} shows the growth of the number of evaluations with problem size for onemax-zeromax. The results indicate that clustering in the objective space is necessary for scalable solution of onemax-zeromax. Furthermore, the results show that here RTR based on nondominated crowding performs better than the elitist replacement of NSGA-II. Finally, the results indicate that UMDA with RTR provides low-order polynomial solution to onemax-zeromax.

\begin{figure}[t]
\centering
\epsfig{file=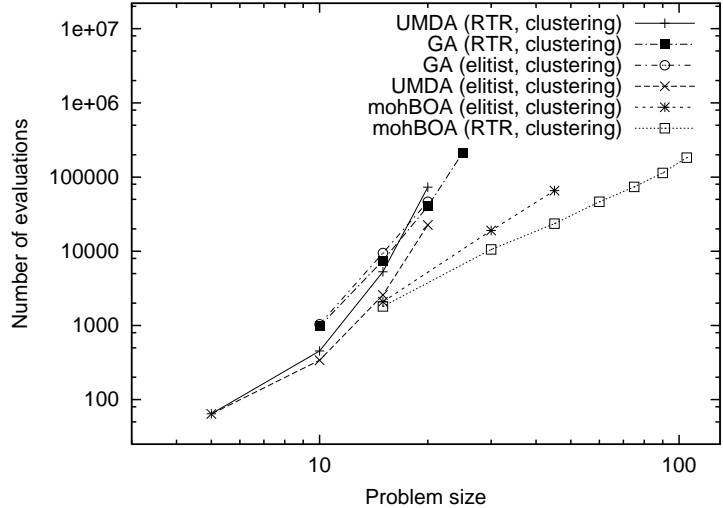,width=0.6\textwidth}
\caption{Results on trap5-invtrap5 indicate that for some decomposable problems it is necessary to both include clustering as well as identify and exploit interactions between interacting string positions or decision variables. Furthermore, they show that RTR performs better than the elitist replacement of NSGA-II and that mohBOA with RTR is capable of solving trap5-invtrap5 in low-order polynomial time.}
\label{fig-results-trap5}
\end{figure}

Figure~\ref{fig-results-trap5} shows the results on trap5-invtrap5. The results show that as expected, trap5-invtrap5 necessitates not only clustering in the objective space like onemax-zeromax but also effective identification and exploitation of interactions between different problem variables also called linkage learning. That is why standard crossover and UMDA fail to solve this problem efficiently and become intractable already for relatively small problems. The algorithm mohBOA with RTR and clustering in the objective space provides best performance and scales up polynomially with problem size. Again, RTR leads to better performance than elitism.

\begin{figure}[t]
\centering
\epsfig{file=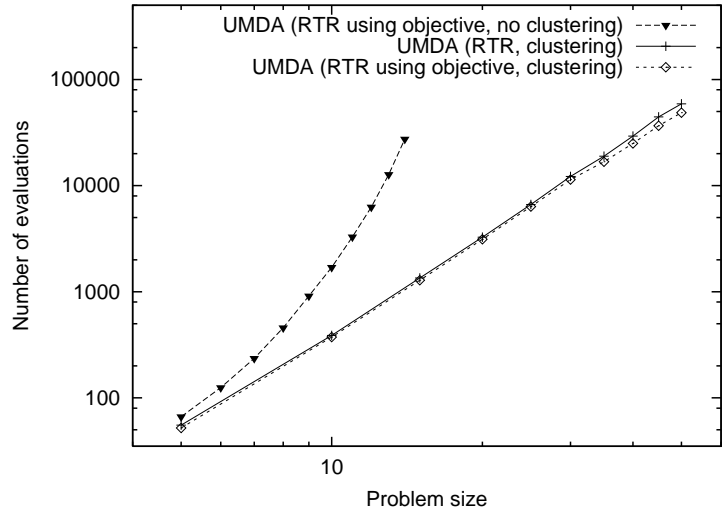, width=0.6\textwidth}
\caption{The results of multiobjective UMDA with RTR using a distance metric in the objective space on onemax-zeromax indicate that clustering in the objective space cannot be replaced with this variant of RTR and that the choice of metric in RTR does not significantly affect performance.}
\label{fig-results-objectiveRTR}
\end{figure}

Figure~\ref{fig-results-objectiveRTR} shows that the performance of UMDA on onemax-zeromax does not change much if the objective space is used to compute the distance of solutions in RTR (as opposed to using standard distance metrics for binary strings). However, using RTR in the objective space is still not capable of ensuring scalable performance if clustering in the objective space is not used indicating that it is insufficient to incorporate niching via replacement based on the distribution of solutions no matter whether the niching method is based on the candidate solutions themselves or their objective values.

\section{Summary and conclusions}
\label{section-conclusions}
This paper discussed scalable optimization of multiobjective decomposable problems where the objectives compete in different partitions of the problem decomposition. The multiobjective hierarchical BOA (mohBOA) was proposed that combines the hierarchical BOA with the nondominated crowding of NSGA-II and clustering in the objective space. By combining one of the most powerful genetic and evolutionary algorithms with NSGA-II and clustering, a scalable multiobjective optimization algorithm for decomposable problems was created. Only problems with two objectives were used in the experiments but the conclusions drawn should apply to problems with more than two objectives.

Experimental results indicate that clustering in the objective space is necessary for scalable optimization of decomposable multiobjective problems. Restricted tournament replacement (RTR) based on nondominated crowding appears to perform better than the elitist replacement used in NSGA-II. Furthermore, to solve arbitrary multiobjective decomposable problems, linkage learning must be considered to effectively identify and process different subproblems. The experiments indicate that mohBOA can solve decomposable multiobjective problems in low-order polynomial time, whereas other compared algorithms fail to scale up well and become intractable for already relatively small problems.


\section*{Acknowledgments}
This work was partially supported by the Research Award and the Research Board at the University of Missouri. Some experiments were done using the hBOA software developed by Martin Pelikan and David E. Goldberg at the University of Illinois at Urbana-Champaign. 
This work was also sponsored by the Air Force Office of Scientific
Research, Air Force Materiel Command, USAF, under grant
F49620-03-1-0129, the National Science Foundation under ITR grant
DMR-99-76550 (at Materials Computation Center), and ITR grant
DMR-0121695 (at CPSD), and the Dept. of Energy under grant
DEFG02-91ER45439 (at Fredrick Seitz MRL). The U.S.  Government is
authorized to reproduce and distribute reprints for government
purposes notwithstanding any copyright notation thereon.

\bibliographystyle{apa-uiuc}
\bibliography{mybib}

\end{document}